\documentclass{article} 
\usepackage[final]{colm2025_conference}

\usepackage{microtype}
\usepackage{amsmath}
\usepackage{hyperref}
\usepackage{url}
\usepackage{booktabs}
\usepackage{graphicx}
\usepackage{multirow}
\usepackage{lineno}
\usepackage{tabularx}
\usepackage{xcolor}
\usepackage{colortbl}
\usepackage{subcaption}
\usepackage{wrapfig}

\definecolor{darkblue}{rgb}{0, 0, 0.5}
\hypersetup{colorlinks=true, citecolor=darkblue, linkcolor=darkblue, urlcolor=darkblue}

\title{VLM-C4L: Continual Core Dataset Learning with Corner Case Optimization via Vision-Language Models for Autonomous Driving}

\author{
Haibo Hu$^{1}$, 
JiaCheng Zuo$^{2}$\thanks{Equal contribution}, 
Lou Yang$^{1}$, 
Yufei Cui$^3$, 
Jianping Wang$^{1}$, 
Nan Guan$^{1}$,\\
\textbf{Jin Wang$^2$}, 
\textbf{Yung-Hui Li$^4$}, 
\textbf{Chun Jason Xue$^5$}\\
$^1$City University of Hong Kong \quad
$^2$Soochow University \quad
$^3$McGill University \\
$^4$Hon Hai Research Institute \quad
$^5$Mohamed bin Zayed University of Artificial Intelligence\\
\texttt{\{haibohu2-c, yanglou3-c\}@my.cityu.edu.hk, \{jianwang, nanguan\}@cityu.edu.hk,}\\
\texttt{jczuo102137@stu.suda.edu.cn, wjin1985@suda.edu.cn, yufeicui92@gmail.com,}\\
\texttt{yunghui.li@foxconn.com, jason.xue@mbzuai.ac.ae}
}

\begin{document}

\ifcolmsubmission
\linenumbers
\fi

\maketitle
\begin{abstract}
With the widespread adoption and deployment of autonomous driving, handling complex environments has become an unavoidable challenge. Due to the scarcity and diversity of extreme scenario datasets, current autonomous driving models struggle to effectively manage corner cases. This limitation poses a significant safety risk—according to the National Highway Traffic Safety Administration (NHTSA), autonomous vehicle systems have been involved in hundreds of reported crashes annually in the United States, occurred in corner cases like sun glare and fog, which caused a few fatal accident.~\cite{tesla2024}. Furthermore, in order to consistently maintain a robust and reliable autonomous driving system, it is essential for models not only to perform well on routine scenarios but also to adapt to newly emerging scenarios—especially those corner cases that deviate from the norm. This requires a learning mechanism that incrementally integrates new knowledge without degrading previously acquired capabilities. However, to the best of our knowledge, no existing continual learning methods have been proposed to ensure consistent and scalable corner case learning in autonomous driving.  To address these limitations, we propose VLM-C4L, a continual learning framework that introduces Vision-Language Models (VLMs) to dynamically optimize and enhance corner case datasets, and VLM-C4L combines VLM-guided high-quality data extraction with a core data replay strategy, enabling the model to incrementally learn from diverse corner cases while preserving performance on previously routine scenarios, thus ensuring long-term stability and adaptability in real-world autonomous driving. We evaluate VLM-C4L on large-scale real-world autonomous driving datasets, including Waymo and the corner case dataset CODA. To assess the effectiveness of our approach, we employ Sparse R-CNN, the strongest model in the CODA benchmark, and Cascade-DETR, a widely recognized model. Experimental results show that VLM-C4L greatly improves object detection in challenging scenarios like light pollution and fog, with AP and AR scores close to those in normal conditions. our code and datasets are here:~\url{www.vlmc4l.site/}
\end{abstract}

\section{Introduction} 
Deep neural network-based autonomous driving technologies have advanced rapidly, achieving significant milestones in real-world deployment, especially in perception tasks such as object detection~\cite{cai2018cascade,jiang2020sp,zhu2020deformable,sun2021sparse}.
These detection models excel at detecting common traffic participants (e.g. cars, pedestrians, and bicycles), primarily due to the availability of extensive and well-curated regular datasets, such as Waymo \cite{waymo}, nuScenes \cite{nuscenes}, and KITTI \cite{kitti}. However, handling corner cases—rare, unusual, and challenging driving conditions—remains a critical bottleneck for achieving robust performance in complex and dynamic environments \cite{zhao2024autonomous,chen2024end}.According to a report released by the U.S. National Highway Traffic Safety Administration (NHTSA), nearly 400 collisions involving Level 2 Advanced Driver Assistance Systems (ADAS) were reported by automakers such as Tesla between July 2023 and May 2024. A significant portion of these incidents resulted from the failure of autonomous driving systems to handle rare or unexpected driving conditions, with Tesla reporting multiple crashes caused by factors such as sun glare, fog, and airborne dust~\cite{tesla2024}.  These real-world failure cases have motivated increasing research attention toward evaluating model robustness under challenging scenarios.

Table~\ref{table:the performance of CODA} presents the object detection metrics(AP/AR) for the CODA dataset~\cite{coda123}, where the models were trained on the Waymo dataset. "WAYMO" represents their object detection performance on the Waymo dataset, while "COMMON" refers to their detection results for common objects (vehicles, pedestrians, and bicycles) on the CODA dataset. "NOVAL" indicates the model's detection performance on corner cases within the CODA dataset, and "CORNER"  combines results from both COMMON and NOVAL. The results clearly demonstrate that models trained on regular datasets struggle to handle corner cases effectively, with most scores falling below 10\%. 
\begin{table}[!tb]
    \centering
    \caption{The Average Precision(AP) and Average Recall(AR) of Models for Autonomous Driving in Waymo and CODA Datasets}
    \label{table:the performance of CODA}
    \renewcommand{\arraystretch}{1.1} 
    \setlength{\tabcolsep}{4pt}
    \small
    \begin{tabular}{c|c|c|c|c|c}
        \hline
        \multirow{2}{*}{\small Method} & \multicolumn{2}{c|}{\small \small WAYMO} & \multicolumn{1}{c|}{\small CORNER} & \multicolumn{1}{c|}{\small COMMON} & \multicolumn{1}{c}{\small NOVEL}\\
        \cline{2-6}
        & \small AP(\%) & \small AR(\%) & \small mAR(\%) & \small mAR(\%) & \small mAR(\%) \\
        \hline
        \small RetinaNet \cite{model1}  & 39.7 & 47.7 & 8.4 & 24.5 & 6.7 \\
        \hline
        \small Faster R-CNN \cite{model2}  & 40.9 & 47.0 & 6.8 & 20.9 & 5.5 \\
        \hline
        \small Cascade R-CNN \cite{model3}& 42.6 & 48.1 & 6.6 & 18.9 & 5.3 \\
        \hline
        \small D-DETR \cite{model4}& 40.4 & 49.8 & 7.3 & 28.5 & 5.2 \\
        \hline
        \small{Sparse R-CNN \cite{sun2021sparse}}  & 38.8 & 49.8 & 10.1 & 29.5 & 7.6 \\
        \hline
        \small Cascade Swin \cite{model5}& 44.2 & 49.0 & 5.4 & 21.8 & 4.3 \\
        \hline
    \end{tabular}
\vspace{-20pt}
\end{table}

To address corner cases, existing autonomous driving models often rely on specialized datasets like CODA and fine-tune on unclassified corner case data~\cite{ce.oda_track1,coda_track2}. While this improves recognition to some extent, it fails to explicitly identify which cases are resolved, which is inadequate for safety-critical applications. Simulators (e.g., CornerSim~\cite{cornersim123}) offer another option, but the Sim-to-Real gap remains a major challenge~\cite{Sim2Real,ralad}. Ultimately, real-world data is essential. However, corner case data is both scarce and highly diverse—for instance, CODA contains fewer than 10,000 scenes, but covers over 40 different situations, compared to 230,000 in Waymo~\cite{waymo} or millions in ONCE~\cite{mao2021one}. Enabling models to handle specific, diverse corner cases with limited data is therefore a key challenge.

Therefore, enabling autonomous driving models to effectively handle specific corner case scenarios while ensuring adaptability to the diversity of corner cases, all within the constraints of limited data, remains an open challenge. Continual learning offers a promising direction to address this issue by allowing models to incrementally adapt to newly emerging corner cases without forgetting previously learned knowledge, thus enhancing long-term generalization across diverse scenarios~\cite{cl4survey,rain}. Moreover, the effectiveness of continual learning relies on the availability of well-defined and representative corner case data, which remains scarce and often ambiguous. Recent breakthroughs in vision-language models (VLMs) have demonstrated their exceptional capabilities in reasoning \cite{vlmReason}, classification \cite{vlmClass,vlmClass2}, and feature extraction \cite{vlmFeature} across various domains. While VLMs have been extensively used in these areas, their potential to address challenges in autonomous driving—particularly in handling corner cases—remains largely unexplored. The logical reasoning and robust memory capabilities of VLMs offer promising opportunities to identify, interpret, and organize corner case scenarios, mitigating the limitations of manually curated datasets and providing a richer foundation for continual learning in autonomous systems.

Based on our findings, we propose VLM-C4L, a continual learning framework that optimizes corner case datasets using VLMs to update core datasets dynamically. VLM-C4L first leverages VLMs to efficiently extract, distribute, and enhance corner case data, forming targeted corner case datasets. It then introduces a tailored continual learning strategy that preserves the original model’s capabilities using core datasets while fine-tuning the model with corner case data to improve its ability to handle corner case tasks. Finally, previously learned corner case scenarios are processed through uncertainty-based core data extraction, updating the core dataset to ensure that knowledge from prior corner cases is retained when adapting to new ones.

In summary, our work presents three major contributions to autonomous driving for corner case:
\begin{itemize}
\item VLM-C4L introduces a novel VLM-based framework for optimizing corner case datasets by systematically extracting, classifying, and enhancing critical samples. Unlike existing methods that fine-tune on unstructured data, VLM-C4L improves model performance through refined, balanced corner case selection. This enables more targeted training and evaluation for future autonomous driving systems.
\item The framework incorporates a continual learning strategy based on core datasets, allowing the model to incrementally learn new corner case scenarios while preserving previously acquired knowledge. This approach mitigates catastrophic forgetting and improves the model’s ability to generalize across diverse and evolving corner cases.
\item We conduct comprehensive experiments on multiple autonomous driving models, demonstrating that VLM-C4L consistently improves performance across various corner case scenarios. The results validate the effectiveness and generalizability of our approach.
\end{itemize}

\section{Related Work}
\subsection{Corner Case Data In Autonomous Driving} 
In autonomous driving, corner cases refer to rare, unexpected scenarios that challenge perception and decision-making systems \cite{cc4identi,cc4identi4a}. These include unusual objects, conditions, or behaviors not typically encountered during standard driving. Visual perception plays a key role in identifying such scenarios. Approaches to address corner cases include human-in-the-loop testing \cite{cc4carla} and simulators like CARLA for generating corner case datasets \cite{cornersim123}. However, transferring these methods to models trained on real-world data remains difficult due to the sim-to-real gap \cite{Sim2Real,ralad}. Real-world datasets like CODA \cite{coda123} focus mainly on evaluation rather than resolution. Some methods leverage decision-making frameworks such as Markov decision processes and deep reinforcement learning \cite{cc4make} to generate and handle corner cases. Despite progress, challenges remain due to the scarcity and variability of corner case data, highlighting the need for more systematic solutions.

\subsection{Vision Language Models In Autonomous Driving}
As autonomous driving advances, there's a growing need for models that not only perceive but also interpret and adapt to complex environments. Recent research highlights the promise of Vision Language Models (VLMs) in enhancing interpretability and performance, especially in human interaction and scene understanding \cite{vlm4human,vlm4planning}. VLMs have been used as teacher models to improve training~\cite{vlm4techer}, regulate driving behavior~\cite{vlm4codriver}, and support planning in complex scenarios~\cite{vlm4planning}. They also show potential in anomaly detection, intent prediction, dynamic traffic analysis, and fine-grained recognition. However, their application to corner case datasets remains unexplored—an opportunity that could greatly improve autonomous systems' handling of rare and challenging situations.

\subsection{Continual Learning}
Continual learning refers to a model’s ability to learn from new data while retaining previously acquired knowledge without experiencing catastrophic forgetting \cite{cl4survey,cl4survey2}. To mitigate this issue, three primary categories of approaches have been developed in computer vision. Regularization-based methods, such as EWC \cite{cl4jj1} and MAS \cite{mas}, constrain parameter updates to preserve past knowledge, but often yield suboptimal results in complex real-world scenarios. Architecture-based methods, including PNN~\cite{pnn} and PackNet \cite{packnet}, modify or expand the model architecture to accommodate new tasks, yet they face scalability limitations, particularly in multi-task learning contexts beyond simple classification. In contrast, replay-based methods—such as DER~\cite{dar} and GEM~\cite{gem}—have emerged as the most widely adopted solutions due to their ability to reinforce memory by selectively revisiting previously seen data . Considering the diversity and sparsity of corner cases in autonomous driving, we adopt a replay-based continual learning paradigm with core data selection as our primary strategy~\cite{rain}. The next section will explain in detail why this method was chosen.

\section{Methodology}
\subsection{Motivation}
Continual learning methods are typically categorized into parameter-isolation, regularization-based, and replay-based approaches. Parameter-isolation methods (e.g., PackNet, HAT) mitigate forgetting but scale poorly—our tests show that handling multiple corner cases can increase model size by over 1.5×. Regularization-based methods (e.g., EWC, RWalk) avoid storing past data by penalizing changes to important parameters, but they require a full pass over the entire dataset after each task to compute statistics like the Fisher Information Matrix. For large-scale datasets like Waymo, this results in long training times (e.g., 48–72 hours on 8 MXC500 GPUs), making them unsuitable for rapid deployment. In contrast, replay-based methods, especially those using core datasets, offer a lightweight and efficient alternative. Traditional replay methods, however, struggle in autonomous driving due to large input sizes and inefficient memory usage. By selecting a small set of representative, discriminative, and diverse samples, we achieve strong performance using less than 1\% of the data—e.g., just 10,000 core Waymo samples. Considering scalability, resource cost, and generalization to corner cases, we adopt a core dataset replay-based framework for continual learning.

\subsection{Problem Formulation}
The core problem we address is how to continually update the core dataset to both preserve the model’s existing capabilities and progressively enhance its parameters for handling new corner case scenarios. This involves maintaining model stability across tasks while enabling efficient and scalable learning from newly emerging, diverse corner cases. Formally, we define our iterative corner case learning process as:
\begin{equation}
    \theta^{(t+1)}, D_{\text{core}}^{(t+1)} = \mathcal{G} \left(D_{\text{core}}^{(t)}, \mathcal{O}_{\text{VLM}}(D_{\text{corner}}^{(t)}), \theta^{(t)} \right),
    \label{eq:problem_formulation}
\end{equation}
where: \( D_{\text{core}}^{(t)} \) is the core dataset at iteration \( t \), containing the most informative samples to maintain model stability. \( D_{\text{corner}}^{(t)} \) represents the available corner case dataset at iteration \( t \). \( \mathcal{O}_{\text{VLM}}(\cdot) \) extracts and enhances meaningful corner case samples. \( \theta^{(t)} \) is the model parameter set at iteration \( t \). \( \mathcal{G}(\cdot) \) represents the continual learning process, which updates the model parameters and expands the core dataset to ensure better recognition of corner cases in future iterations.
\subsection{Model Optimization with Core and Corner Case Data}
To enhance model robustness in complex driving scenarios, we optimize the autonomous driving model by integrating the dynamically updated core dataset and the refined corner case dataset. The training process is defined as:
\begin{equation}
    \theta^{(t+1)} = \arg\min_{\theta} \sum_{(x,y) \in D_{\text{core}}^{(t)} \cup D_c^{(t)}} \mathcal{L}(f_{\theta}(x), y),
    \label{eq:model_training}
\end{equation}
where \( f_{\theta}(x) \) represents the model's prediction, and \( \mathcal{L}(\cdot) \) is the task-specific loss. The core dataset \( D_{\text{core}}^{(t)} \) preserves essential knowledge, while \( D_c^{(t)} \) consists of newly extracted corner case samples.

A continual learning strategy balances knowledge retention and adaptation, preventing catastrophic forgetting while ensuring progressive improvement. The updated model parameters \( \theta^{(t+1)} \) incorporate new corner case information while maintaining generalization to diverse driving conditions.

\begin{figure*}
\centering
\includegraphics[width=1\textwidth]{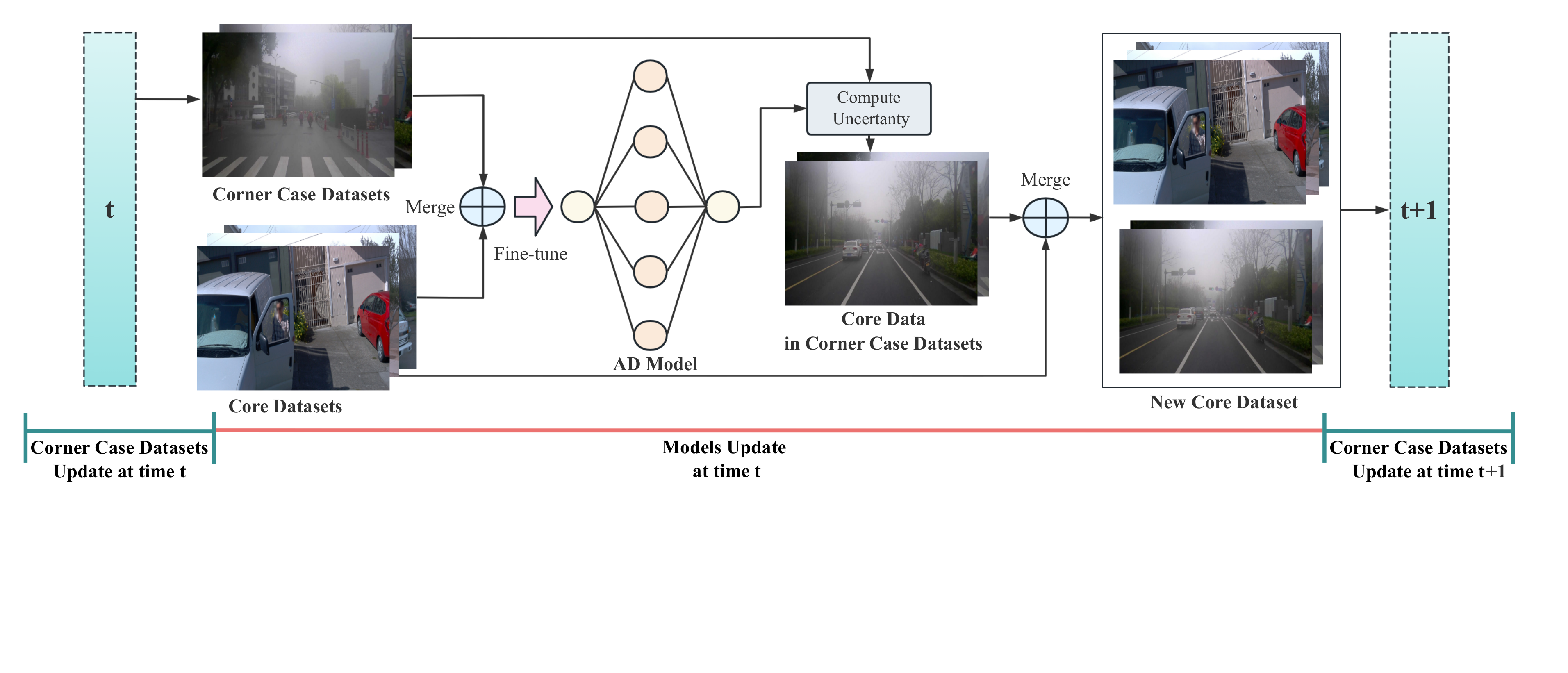}
\vspace{-60pt}
\caption{Continual Update of the AD Model under the VLM-C4L Framework}
\vspace{2pt}
\label{fig:continual}
\vspace{-10pt}
\end{figure*}
\subsection{Continual Core Dataset Update}
A crucial aspect of our approach is the continual update of the core dataset to incorporate new corner cases while retaining critical past knowledge. We use an uncertainty-based selection mechanism to update the core dataset:
\begin{equation}
D_{\text{core}}^{(t+1)} = D_{\text{core}}^{(t)} \cup \mathcal{U} (D_{\text{c}}^{(t)}, f_{\theta^{(t+1)}}),
\end{equation}
where $\mathcal{U} (D_{\text{c}}^{(t)}, f_{\theta^{(t+1)}})$ selects the most informative corner case samples based on an uncertainty criterion derived from stochastic perturbation-based prediction consistency. Specifically, the uncertainty of a sample $x$ is computed by applying multiple data augmentations and measuring the consistency of model predictions across these perturbed versions. The selection mechanism favors samples with high prediction variance, indicating uncertain decisions by the model.

The uncertainty-based selection function $\mathcal{U}$ is defined as:
\begin{equation}
\mathcal{U}(D_{\text{c}}^{(t)}, f_{\theta^{(t+1)}}) = \{ x \mid \sigma_x > \tau, x \in D_{\text{c}}^{(t)} \},
\end{equation}
where $\sigma_x$ represents the prediction variance of sample $x$, computed as:
\begin{equation}
\sigma_x = 1 - \frac{\max_c \sum_{i=1}^{N} \mathbb{I} (\arg\max f_{\theta^{(t+1)}} (\tilde{x}_i) = c )}{N},
\end{equation}
where $\tilde{x}_i$ represents the $i$-th augmented version of $x$, $N$ is the number of augmentations, and $\mathbb{I}(\cdot)$ is the indicator function that counts occurrences of the most frequent predicted class $c$. The threshold $\tau$ determines the level of uncertainty required for a sample to be added to the updated core dataset.

Figure~\ref{fig:continual} illustrates the continual core dataset learning framework. The uncertainty-based selection module evaluates the informativeness of each sample, filtering and merging high-value corner cases into the new core dataset. This continual learning cycle ensures that the model progressively adapts to previously unseen scenarios while retaining critical knowledge from past training.

\subsection{Corner Case Data Extraction and Optimization}
Identifying corner cases in autonomous driving has traditionally relied on manual inspection and rule-based filtering, which is time-consuming, inconsistent, and semantically limited. To overcome these challenges, we propose a Vision-Language Model (VLM)-based pipeline that automatically extracts, partitions, and augments corner case data with minimal human effort. This improves efficiency, consistency, and coverage while reducing manual overhead. Our goal is to extract specific corner cases, ensure balanced data through mean partitioning, and apply augmentation to enhance model robustness. The process is defined as follows:
\begin{equation}
    \mathcal{O}_{\text{VLM}}(D_{\text{corner}}^{(t)}) = 
    \mathcal{A} \left( 
        \mathcal{M} \left( 
            \mathcal{E}_{\text{VLM}} \left( D_{\text{corner}}^{(t)} \right) 
        \right) 
    \right)
    \quad
    \substack{
        \mathcal{M}(*^{(t)}) = \left( D_{\text{train}}^{(t)}, D_{\text{val}}^{(t)}, D_{\text{test}}^{(t)} \right) \\
        \text{(Mean partitioning function)}
    }
    \label{eq:corner_case_processing}
\end{equation}
\begin{figure*}
\centering
\includegraphics[width=1\textwidth]{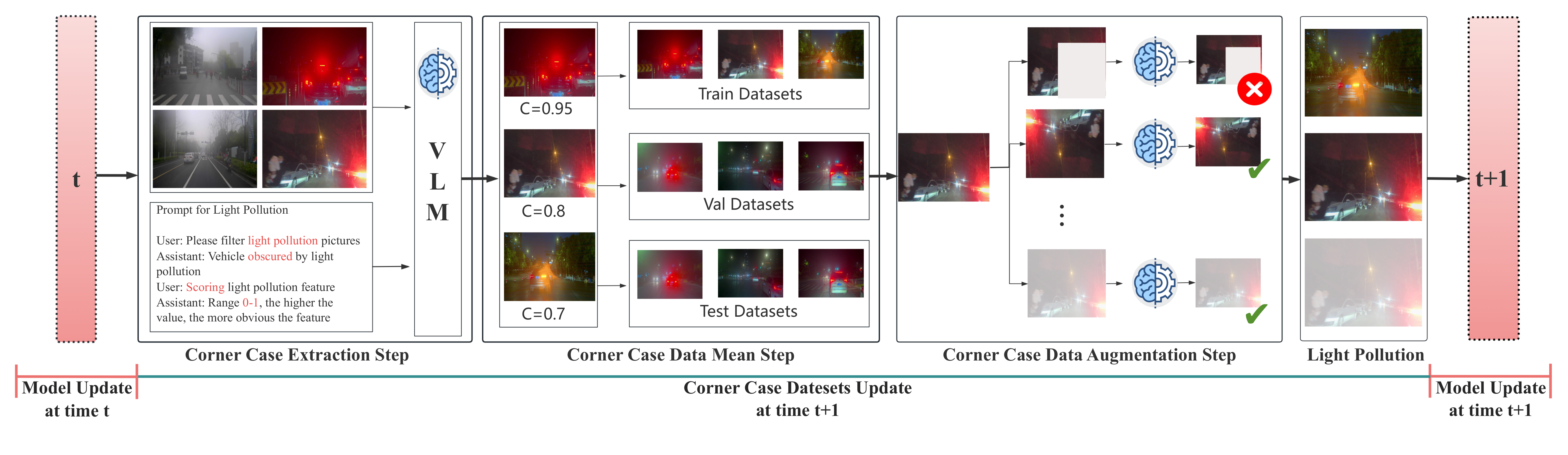}
\vspace{-20pt}
\caption{Continual Update of Corner Case Dataset under the VLM-C4L Framework}
\vspace{2pt}
\label{fig:overview}
\vspace{-10pt}
\end{figure*}
\begin{wrapfigure}{l}{0.5\textwidth}
  \centering
  \includegraphics[width=0.48\textwidth]{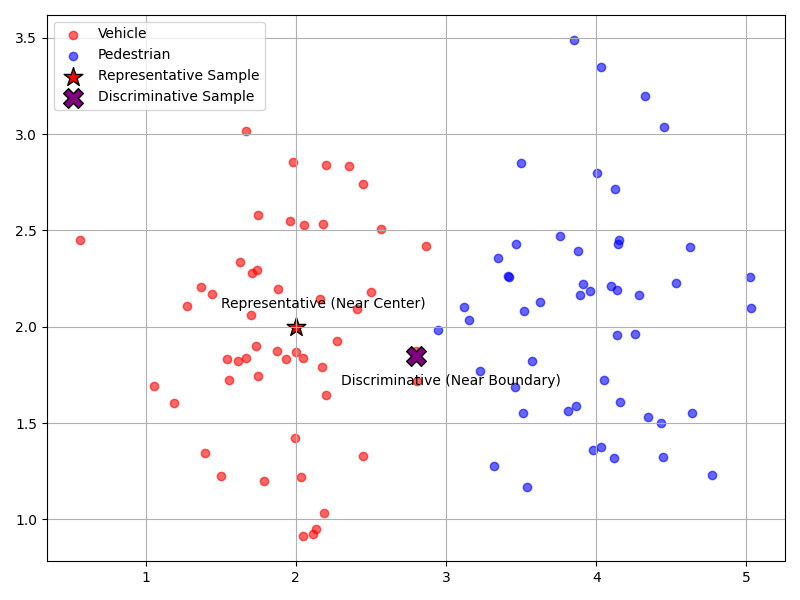}
  \caption{Representative and Discriminative Samples in Feature Space}
  \vspace{-10pt}
  \label{fig:feature}
\end{wrapfigure}
The function \( \mathcal{E}_{\text{VLM}} \) utilizes the Vision-Language Model (VLM) to identify and extract relevant corner cases from the dataset based on semantic understanding, forming a refined subset \( D_{\text{corner}}^{(t)} \). The extracted dataset is then processed by \( \mathcal{M} \), which applies a mean partitioning strategy to ensure balanced data distribution, splitting it into training, validation, and test sets. This step is crucial for maintaining a dataset that facilitates effective model training. As illustrated in Figure~\ref{fig:feature}, the dataset contains both \textit{representative samples}, which lie near the center of the class distribution, and \textit{discriminative samples}, which lie closer to the decision boundary. To ensure robust learning, it is essential that both types of samples are evenly distributed across the train/val/test splits. Representative samples help the model capture the general characteristics of a class, while discriminative samples are critical for fine-tuning the model's decision-making ability near ambiguous regions. Uniform inclusion of both types enhances the model’s generalization and stability in downstream tasks. Finally, \( \mathcal{A} \) performs data augmentation, addressing the issue of insufficient supplementary data by applying transformations such as rotation, scaling, and flipping to enhance data variability and robustness. This ensures that the model has sufficient and diverse data for learning, ultimately improving its ability to generalize to unseen corner case scenarios.

Figure~\ref{fig:overview} illustrates the VLM-guided corner case optimization process. In the Corner Case Extraction step, the VLM is prompted with descriptions of challenging scenarios (e.g., light pollution and foggy) to retrieve and categorize relevant samples. Each sample receives a confidence score \( C \), indicating its degree of corner case feature relevance. The Mean Partitioning step distributes both high- and low-confidence samples evenly across training, validation, and test sets—ensuring that the model learns both representative and boundary-critical features while avoiding data bias. Finally, in the Data Augmentation step, transformations such as blurring, brightness adjustment, and geometric perturbation enhance variability. The resulting dataset is fed into the continual learning process, enabling progressive adaptation to diverse corner cases while preserving generalization.

\section{Experiment}
We evaluated the effectiveness of the proposed VLM-C4L method using images from the large-scale, real-world autonomous driving datasets CODA~\cite{coda123} and WAYMO~\cite{waymo}, enabling a comprehensive assessment of its performance. This section presents the dataset details, implementations and VLM settings, followed by the experimental results and outcomes of the ablation studies.

\subsection{Implementation Details}
{\bf Dataset:} Our experiments utilize three distinct dataset splits—WAYMO, Light Pollution, and Fog—to evaluate the effectiveness of VLM-C4L. These splits are derived from two well-known datasets: WAYMO and CODA. 
\textit{WAYMO} split is sourced from the Waymo validation split, which contains 199,938 images captured by the vehicle’s five 360-degree panoramic cameras. The dataset includes annotations for three primary categories: vehicles, pedestrians, and cyclists. WAYMO serves as a baseline for general object detection tasks in autonomous driving. 

{\bf Corner Case:} Due to limited hardware resources, we prioritize glare and fog as our primary corner cases, as these conditions are identified as major contributing factors in autonomous driving failures according to the NHTSA report. we utilize VLMs to extract corner case datasets from the CODA dataset, resulting in 1,202 images for light pollution scenarios and 2,509 images for fog scenarios. To ensure quality, we conducted a thorough manual comparison and will release the refined datasets along with the code for reproducibility. In the end, the datasets are divided into training, validation, and test sets in a 4:3:3 ratio.

{\bf Models:} We utilized the Sparse R-CNN and Cascade-DETR as baseline detectors, following their official implementations and original training parameters without modifications. Due to the domain gap between COCO and the datasets used in our experiments (Waymo and CODA), detectors were pre-trained on the Waymo training set to ensure fair evaluation. Fine-tuning with VLM-C4L involved freezing the backbone of Sparse R-CNN and the encoder of Cascade-DETR. All training and evaluation were conducted on a workstation equipped with 8 MXC500 GPU cards.

{\bf Metrics:} We follow the CODA dataset evaluation pipeline, using Average Precision (AP) and Average Recall (AR) as the evaluation metrics. AP measures the precision of detected objects at different recall thresholds, while AR quantifies the model’s ability to identify objects across all recall levels.

{\bf VLM Setting:} In our experiments, we utilized the pre-trained CLIP ViT-B/32 model with 3.2 billion parameters, offering substantial learning capacity and well-suited for medium-scale datasets like CODA. The model maps images and our predefined keywords into a shared feature space, enabling it to calculate similarities and precisely filter corner cases. Additionally, uncertainty assessments conducted on these datasets further enhance the accuracy of experimental results and the model’s overall effectiveness.

\subsection{Experimental Results}
We evaluated the effectiveness of VLM-C4L on two datasets: the Waymo dataset and corner case datasets selected from CODA. Two widely recognized 2D object detection models, Sparse R-CNN and Cascade-DETR, were used as baselines. Sparse R-CNN, originally proposed as the baseline model in CODA, and Cascade-DETR, a more recent model, were tested in both their original configurations and after fine-tuning with VLM-C4L. All experiments were conducted in a consistent hardware and software environment to ensure fairness and reliability. Table~\ref{table:Comparative experiment} presents the experimental metrics, comparing the performance of the two models on the Waymo dataset (WAYMO) and two corner case scenarios (Light Pollution and Fog) with and without VLM-C4L.

{\bf Sparse R-CNN Results:} the Sparse R-CNN model fine-tuned with VLM-C4L exhibited a slight decrease in performance on the WAYMO dataset, with AP decreasing by 1.7\% and AR by 0.8\%. However, significant improvements were observed on the corner case scenarios. On Light Pollution, AP increased from 14.6\% to 32.5\% and $AR_{100}$ from 30.2\% to 52.3\%. Similarly, on Fog, AP improved from 16.7\% to 30.7\% and $AR_{100}$ from 29.8\% to 43.5\%. These results highlight the effectiveness of VLM-C4L in addressing domain class incremental issues and improving model performance under challenging scenarios.
\begin{table*}[!tb]
    \centering
    \caption{Comparison of Object Detection Performance with VLM-C4L \textit{\tiny{(AP, AR in \%)}}}
    \label{table:Comparative experiment}
    \renewcommand{\arraystretch}{1.0} 
    \setlength{\tabcolsep}{0.5pt} 
    \small
    \begin{tabular}{c|c|c|c|c|c|c|c|c|c|c|c|c|c|cc}
        \hline
        \multirow{2}{*}{Methods}  & \multicolumn{2}{c|}{WAYMO} & \multicolumn{6}{c|}{Light Pollution} &\multicolumn{6}{c}{Foggy} \\ 
        \cline{2-15}
        & AP & AR & AP& AP\textsubscript{50}& AP\textsubscript{75} & AR\textsubscript{1} & AR\textsubscript{10} & AR\textsubscript{100} & AP& AP\textsubscript{50}& AP\textsubscript{75}& AR\textsubscript{1} & AR\textsubscript{10} & AR\textsubscript{100}\\
        \hline
        \tiny{Sparse R-CNN} & \cellcolor{blue!20}\textbf{36.4} & \cellcolor{blue!20}\textbf{45.9} & 14.6 & 25.5 & 15.1 & 12.0 & 26.9 & 30.2 & 16.7 & 28.0 & 17.8 & 11.9 & 27.3 & 29.8 \\  \tiny{Sparse R-CNN+\textbf{VLM-C4L(ours)}} & 34.7 & 45.1 & \cellcolor{blue!20}\textbf{32.5}$\uparrow$ & \cellcolor{blue!20}\textbf{52.4}$\uparrow$ & \cellcolor{blue!20}\textbf{34.1}$\uparrow$ & \cellcolor{blue!20}\textbf{20.6}$\uparrow$ & \cellcolor{blue!20}\textbf{49.0}$\uparrow$ & \cellcolor{blue!20}\textbf{52.3}$\uparrow$ & \cellcolor{blue!20}\textbf{30.7}$\uparrow$ & \cellcolor{blue!20}\textbf{46.5}$\uparrow$ & \cellcolor{blue!20}\textbf{33.2}$\uparrow$ & \cellcolor{blue!20}\textbf{17.6}$\uparrow$ & \cellcolor{blue!20}\textbf{41.0}$\uparrow$ & \cellcolor{blue!20}\textbf{43.5}$\uparrow$\\
        \cline{1-1}\cline{2-15}
        \tiny{Cascade-DETR} & \cellcolor{blue!20}\textbf{35.5} & \cellcolor{blue!20}\textbf{48.1} & 18.8 & 33.0 & 18.8 & 14.1 & 31.9 & 38.1 & 18.0 & 30.8 & 18.4 & 12.0 & 28.7 & 33.5\\
         \tiny{Cascade-DETR+\textbf{VLM-C4L(ours)}} & 33.3 & 46.5 & \cellcolor{blue!20}\textbf{30.4}$\uparrow$ & \cellcolor{blue!20}\textbf{48.3}$\uparrow$ & \cellcolor{blue!20}\textbf{31.2}$\uparrow$ & \cellcolor{blue!20}\textbf{20.7}$\uparrow$ & \cellcolor{blue!20}\textbf{49.6}$\uparrow$ & \cellcolor{blue!20}\textbf{54.1}$\uparrow$ & \cellcolor{blue!20}\textbf{24.9}$\uparrow$ & \cellcolor{blue!20}\textbf{40.2}$\uparrow$ & \cellcolor{blue!20}\textbf{25.4}$\uparrow$ & \cellcolor{blue!20}\textbf{16.9}$\uparrow$ & \cellcolor{blue!20}\textbf{39.5}$\uparrow$ & \cellcolor{blue!20}\textbf{40.8}$\uparrow$\\
        \hline
    \end{tabular}
\end{table*}
\begin{table*}[!tb]
    \centering
    \caption{Comparison of Object Detection Performance with Different Corner Case Data Subsets \textit{\tiny{(AP, AR in \%)}}}
    \small
    \label{table:ablation_study_combined_single_column}
    \renewcommand{\arraystretch}{1.0} 
    \setlength{\tabcolsep}{0.6pt} 
    \begin{tabular}{c|c|c|c|c|c|c|c|c|c|c|c|c|c|cc}
        \hline
        \multirow{2}{*}{Methods}  & \multicolumn{2}{c|}{WAYMO} & \multicolumn{6}{c|}{Light Pollution} &\multicolumn{6}{c}{Foggy}\\ 
        \cline{2-15}
       & AP & AR & AP & AP\textsubscript{50} & AP\textsubscript{75} & AR\textsubscript{1} & AR\textsubscript{10} & AR\textsubscript{100} & AP & AP\textsubscript{50} & AP\textsubscript{75} & AR\textsubscript{1} & AR\textsubscript{10} & AR\textsubscript{100} \\
        \hline
\tiny{Sparse R-CNN+\textbf{Random}}& 34.6& 43.8 & 20.4 & 31.5 & 21.1 & 15.0 & 35.9 & 39.0& 22.4 & 37.4 & 25.1 & 14.3 & 33.3 & 35.3 \\
\tiny{Sparse R-CNN+\textbf{VLM-DB}} & \cellcolor{blue!20}\textbf{34.9} & \cellcolor{blue!20}\textbf{44.9} & 23.9 & 39.9 & 24.4& 17.2 & 39.8& 43.5 & 23.8 & 37.4 & 26.2& 14.8 & 34.0& 36.4\\
\tiny{Sparse R-CNN+\textbf{Random+Data Aug}}& 34.8 & 44.8  & 29.7 & 52.3& 33.5 & 19.2 & 48.0 & 50.9 & 27.3 & 43.6& 32.5 & 16.2 & 39.8 & 43.1 \\
\tiny{Sparse R-CNN+\textbf{VLM-C4L (ours)}}& 34.7 & 45.1 & \cellcolor{blue!20}\textbf{32.5} & \cellcolor{blue!20}\textbf{52.4} & \cellcolor{blue!20}\textbf{34.1} & \cellcolor{blue!20}\textbf{20.6} & \cellcolor{blue!20}\textbf{49.0} & \cellcolor{blue!20}\textbf{52.3} & \cellcolor{blue!20}\textbf{30.7} & \cellcolor{blue!20}\textbf{46.5} & \cellcolor{blue!20}\textbf{33.2} & \cellcolor{blue!20}\textbf{17.6} & \cellcolor{blue!20}\textbf{41.0} & \cellcolor{blue!20}\textbf{43.5}\\
        \cline{1-1}\cline{2-15}
        
\tiny{Cascade-DETR+\textbf{Random}}& 34.5 & 47.5 & 18.3 & 34.9 & 19.9 & 15.3 & 32.7 & 35.9 & 16.9 & 28.9 & 17.8 & 11.3 & 27.7 & 31.3\\
\tiny{Cascade-DETR+\textbf{VLM-DB}} & \cellcolor{blue!20}\textbf{34.8} & \cellcolor{blue!20}\textbf{47.8} & 21.6 & 36.2 & 22.6 & 15.4 & 37.1 & 44.4 & 19.1 & 31.7 & 20.0 & 12.9 & 30.1 & 36.0\\
\tiny{Cascade-DETR+\textbf{Random+Data Aug} }& 31.5 & 44.8 & 27.6 & 45.3 & 27.7& 18.4 & 47.7& 51.3 & 19.7 & 34.3 & 23.6& 13.4 & 30.7& 33.8\\
\tiny{Cascade-DETR+\textbf{VLM-C4L (ours)}} & 33.3 & 46.5 & \cellcolor{blue!20}\textbf{30.4} & \cellcolor{blue!20}\textbf{48.3} & \cellcolor{blue!20}\textbf{31.2} & \cellcolor{blue!20}\textbf{20.7} & \cellcolor{blue!20}\textbf{49.6} & \cellcolor{blue!20}\textbf{54.1} & \cellcolor{blue!20}\textbf{24.9} & \cellcolor{blue!20}\textbf{40.2} & \cellcolor{blue!20}\textbf{25.4} & \cellcolor{blue!20}\textbf{16.9} & \cellcolor{blue!20}\textbf{39.5} & \cellcolor{blue!20}\textbf{40.8} \\
        \hline
    \end{tabular}
    \vspace{-10pt}
\end{table*}

{\bf Cascade-DETR Results: } the Cascade-DETR fine-tuned with VLM-C4L demonstrated substantial accuracy improvements on the corner case datasets. On Light Pollution, AP increased by 11.6\% and $AR_{100}$ by 16.0\%, while on Fog, AP increased by 6.9\% and AR by 7.3\%. However, a slight decline was observed on the WAYMO dataset, with AP decreasing by 2.2\% and $AR_{100}$ by 1.6\%. These findings indicate that VLM-C4L effectively enhances generalization capabilities and mitigates the issue of catastrophic forgetting.

\subsection{Ablation Study}
To evaluate the effectiveness of VLM-C4L, we conducted a series of ablation studies to examine key factors influencing model performance, including dataset partitioning, confidence thresholds, and core dataset size.

{\bf Effectiveness of VLM-Based Dataset Optimization:} 
To validate the effectiveness of our proposed VLM-C4L method in handling corner case data, we compare it against three baselines: Random, VLM+DB, and Random+Data Aug.

\begin{itemize}
\item \textbf{Random} represents a baseline where corner case data is split randomly into training, validation, and test sets. VLM-C4L significantly outperforms this strategy, with Light Pollution AP increasing from 20.4 to 32.5 and AR from 15.0 to 20.6, and Foggy AP increasing from 22.4 to 30.7 and AR from 14.3 to 17.6. On WAYMO, AP improves slightly from 34.6 to 34.7 and AR from 43.8 to 45.1. These results indicate that VLM-C4L offers a more efficient way to utilize corner case data than random partitioning.
\item \textbf{VLM+DB} employs vision-language models to split data evenly across training, validation, and test sets based on confidence levels, but without applying data augmentation. Compared to Random, this VLM-guided approach already improves performance (e.g., Light Pollution AP from 20.4 to 23.9). However, VLM-C4L further boosts performance with its curriculum-based strategy and data replay, achieving Light Pollution AP of 32.5 and Foggy AP of 30.7. Although VLM+DB achieves the highest AP on WAYMO (34.9 vs. 34.7), the gap is negligible, suggesting that our method provides a better overall balance.
\item \textbf{Random+Data Aug} explores the impact of data augmentation after random data splitting. Compared to Random, this approach improves model performance (e.g., Light Pollution AP from 20.4 to 29.7), confirming the effectiveness of data augmentation. However, VLM-C4L, which combines VLM-guided allocation with augmentation and replay, performs even better—reaching 32.5 AP on Light Pollution and 30.7 AP on Foggy—highlighting that intelligent data selection amplifies the benefit of augmentation.
\end{itemize}

{\bf Impact of Confidence Thresholds on Core Dataset Filtering:} We evaluated the effect of different confidence thresholds (0.2, 0.4, 0.6) on core dataset filtering to remove low-confidence data and assess image uncertainty. As shown in Table~\ref{table:4}, training with a 0.2 threshold slightly improves AP and AR in corner case scenarios (e.g., $+0.6$\% AP in Light Pollution) compared to 0.4, but performs worse on the WAYMO dataset ($-0.3$\% AP, $+0.9$\% AR). Both 0.2 and 0.4 outperform 0.6 across all metrics. Thus, 0.4 offers a better balance between generalization and accuracy and is adopted in our experiments.

{\bf Effect of Core Dataset Size on Performance:} The size of the core dataset significantly affects model performance and forgetting. As shown in Table~\ref{table:5}, using only 3000 core samples leads to performance gains on corner case datasets (e.g., $+0.8$\% AP under light pollution) but causes a notable drop on WAYMO ($-0.8$\% AP, $-1.3$\% AR). Increasing the core set to 20000 improves performance on WAYMO but degrades results on corner cases. A size of 10000 offers a better trade-off, balancing generalization and robustness across datasets.
\begin{table*}[!t]
    \centering
    \caption{Comparison of Object Detection Performance with Different Core Confidence Ratios and Number of Core Data \textit{\tiny{(AP, AR in \%)}}}
    \label{table:comparison}
    \renewcommand{\arraystretch}{1.0} 
    \setlength{\tabcolsep}{1.4pt} 
    \label{table:4}
    \begin{tabular}{c|cc|cccccc|cccccc}
        \hline
        \multirow{2}{*}{Conf. Ratios} & \multicolumn{2}{c|}{WAYMO} & \multicolumn{6}{c|}{\shortstack{Light Pollution}} & \multicolumn{6}{c}{Fog} \\ 
        \cline{2-15}
        & AP & AR & AP & AP\textsubscript{50} & AP\textsubscript{75} & AR\textsubscript{1} & AR\textsubscript{10} & AR\textsubscript{100} &  AP & AP\textsubscript{50} & AP\textsubscript{75} & AR\textsubscript{1} & AR\textsubscript{10} & AR\textsubscript{100} \\
        \hline
        \cline{2-15}
        $\tau = 0.2$ & 34.6 & 44.0 & \cellcolor{blue!20}\textbf{24.5} & \cellcolor{blue!20}\textbf{41.3} & \cellcolor{blue!20}\textbf{26.6} & \cellcolor{blue!20}\textbf{18.1} & \cellcolor{blue!20}\textbf{40.0} & \cellcolor{blue!20}\textbf{43.8} & \cellcolor{blue!20}\textbf{23.8} & \cellcolor{blue!20}\textbf{37.3}& \cellcolor{blue!20}\textbf{26.5} & \cellcolor{blue!20}\textbf{14.9} & \cellcolor{blue!20}\textbf{34.0} & \cellcolor{blue!20}\textbf{36.7}\\
        $\tau=0.4$ & \cellcolor{blue!20}\textbf{34.9} & \cellcolor{blue!20}\textbf{44.9} & 23.9 & 39.9 & 24.4 & 17.2& 39.8 & 43.5 & 23.8 & 37.3 & 26.2 & 14.8 & 34.0 & 36.4  \\
        $\tau=0.6$ & 34.2 & 43.4 & 23.5 & 39.5 & 24.6 & 16.3& 38.4& 41.8 & 23.0 & 35.9& 25.4 & 14.7 & 32.8 & 35.4 \\
        \hline
    \end{tabular}
\end{table*}
\begin{table*}[!t]
    \centering
    \caption{Comparison of Object Detection Performance with Different Core Confidence Ratios \textit{\tiny{(AP, AR in \%)}}}
    \label{table:5}
    \renewcommand{\arraystretch}{1.0} 
    \setlength{\tabcolsep}{1.8pt} 
\begin{tabular}{c|cc|cccccc|ccccccc}
            \hline
            \multirow{2}{*}{Core Data} & \multicolumn{2}{c|}{WAYMO} & \multicolumn{6}{c|}{Light Pollution} & \multicolumn{6}{c}{Fog} \\ 
            \cline{2-15}
            & AP & AR &  AP & AP\textsubscript{50} & AP\textsubscript{75} & AR\textsubscript{1} & AR\textsubscript{10} & AR\textsubscript{100} &  AP & AP\textsubscript{50} & AP\textsubscript{75} & AR\textsubscript{1} & AR\textsubscript{10} & AR\textsubscript{100} \\
            \hline
            $D_{\text{core}}^{(1st)}=3000$ & 34.6 & 43.8 & \cellcolor{blue!20}\textbf{26.8} & \cellcolor{blue!20}\textbf{43.9}& \cellcolor{blue!20}\textbf{29.1} & \cellcolor{blue!20}\textbf{18.3} & \cellcolor{blue!20}\textbf{41.8} & \cellcolor{blue!20}\textbf{45.4}& \cellcolor{blue!20}\textbf{24.3} & \cellcolor{blue!20}\textbf{37.8}& \cellcolor{blue!20}\textbf{26.8} & \cellcolor{blue!20}\textbf{15.1}& \cellcolor{blue!20}\textbf{35.3} & \cellcolor{blue!20}\textbf{38.0} \\
            $D_{\text{core}}^{(1st)}=10000$ & 34.9 & 44.9 & 23.9 & 39.9 & 24.4 & 17.2 & 39.8 & 43.5 & 23.8 & 37.4& 26.2 & 14.8 & 34.0 & 36.4 \\
            $D_{\text{core}}^{(1st)}=20000$ & \cellcolor{blue!20}\textbf{35.4}  & \cellcolor{blue!20}\textbf{44.9} & 19.1 & 32.9 & 19.7 & 15.8 & 33.8 & 36.4 & 20.7 & 33.8& 22.5 & 13.8 & 31.4 & 33.5 \\
            \hline
        \end{tabular}
\vspace{-10pt}
\end{table*}

\section{Conclusion}
We introduce VLM-C4L, a Vision-Language Model driven framework for enhancing autonomous driving models in handling corner cases. By leveraging the semantic and reasoning capabilities of VLMs, our approach optimizes and augments corner case datasets while incorporating continual learning to adapt to diverse and complex scenarios. Experiments on CODA and Waymo demonstrate that VLM-C4L effectively classifies, partitions, and augments data, improving performance in challenging conditions like light pollution and fog. Results with Sparse R-CNN and Cascade-DETR further validate its effectiveness, showing strong potential for improving robustness in real-world autonomous driving.

\bibliography{colm2025_conference}

\begin{thebibliography}{42}
\providecommand{\natexlab}[1]{#1}
\providecommand{\url}[1]{\texttt{#1}}
\expandafter\ifx\csname urlstyle\endcsname\relax
  \providecommand{\doi}[1]{doi: #1}\else
  \providecommand{\doi}{doi: \begingroup \urlstyle{rm}\Url}\fi

\bibitem[Addepalli et~al.(2024)Addepalli, Asokan, Sharma, and Babu]{vlmClass2}
Sravanti Addepalli, Ashish~Ramayee Asokan, Lakshay Sharma, and R~Venkatesh Babu.
\newblock Leveraging vision-language models for improving domain generalization in image classification.
\newblock In \emph{Proceedings of the IEEE/CVF Conference on Computer Vision and Pattern Recognition}, pp.\  23922--23932, 2024.

\bibitem[Aljundi et~al.(2018)Aljundi, Babiloni, Elhoseiny, Rohrbach, and Tuytelaars]{mas}
Rahaf Aljundi, Francesca Babiloni, Mohamed Elhoseiny, Marcus Rohrbach, and Tinne Tuytelaars.
\newblock Memory aware synapses: Learning what (not) to forget.
\newblock In \emph{Proceedings of the European conference on computer vision (ECCV)}, pp.\  139--154, 2018.

\bibitem[Bang et~al.(2021)Bang, Kim, Yoo, Ha, and Choi]{rain}
Jihwan Bang, Heesu Kim, YoungJoon Yoo, Jung-Woo Ha, and Jonghyun Choi.
\newblock Rainbow memory: Continual learning with a memory of diverse samples.
\newblock In \emph{2021 IEEE/CVF Conference on Computer Vision and Pattern Recognition (CVPR)}, Jun 2021.
\newblock \doi{10.1109/cvpr46437.2021.00812}.
\newblock URL \url{http://dx.doi.org/10.1109/cvpr46437.2021.00812}.

\bibitem[Bogdoll et~al.(2021)Bogdoll, Breitenstein, Heidecker, Bieshaar, Sick, Fingscheidt, and Z{\"o}llner]{cc4identi}
Daniel Bogdoll, Jasmin Breitenstein, Florian Heidecker, Maarten Bieshaar, Bernhard Sick, Tim Fingscheidt, and Marius Z{\"o}llner.
\newblock Description of corner cases in automated driving: Goals and challenges.
\newblock In \emph{Proceedings of the IEEE/CVF International Conference on Computer Vision}, pp.\  1023--1028, 2021.

\bibitem[Bolte et~al.(2019)Bolte, Bar, Lipinski, and Fingscheidt]{cc4identi4a}
Jan-Aike Bolte, Andreas Bar, Daniel Lipinski, and Tim Fingscheidt.
\newblock Towards corner case detection for autonomous driving.
\newblock In \emph{2019 IEEE Intelligent vehicles symposium (IV)}, pp.\  438--445. IEEE, 2019.

\bibitem[Buzzega et~al.(2020)Buzzega, Boschini, Porrello, Abati, and Calderara]{dar}
Pietro Buzzega, Matteo Boschini, Angelo Porrello, Davide Abati, and Simone Calderara.
\newblock Dark experience for general continual learning: a strong, simple baseline.
\newblock \emph{Advances in neural information processing systems}, 33:\penalty0 15920--15930, 2020.

\bibitem[Caesar et~al.(2020)Caesar, Bankiti, Lang, Vora, Liong, Xu, Krishnan, Pan, Baldan, and Beijbom]{nuscenes}
Holger Caesar, Varun Bankiti, Alex~H Lang, Sourabh Vora, Venice~Erin Liong, Qiang Xu, Anush Krishnan, Yu~Pan, Giancarlo Baldan, and Oscar Beijbom.
\newblock nuscenes: A multimodal dataset for autonomous driving.
\newblock In \emph{Proceedings of the IEEE/CVF conference on computer vision and pattern recognition}, pp.\  11621--11631, 2020.

\bibitem[Cai \& Vasconcelos(2018{\natexlab{a}})Cai and Vasconcelos]{cai2018cascade}
Zhaowei Cai and Nuno Vasconcelos.
\newblock Cascade r-cnn: Delving into high quality object detection.
\newblock In \emph{Proceedings of the IEEE conference on computer vision and pattern recognition}, pp.\  6154--6162, 2018{\natexlab{a}}.

\bibitem[Cai \& Vasconcelos(2018{\natexlab{b}})Cai and Vasconcelos]{model3}
Zhaowei Cai and Nuno Vasconcelos.
\newblock Cascade r-cnn: Delving into high quality object detection.
\newblock In \emph{Proceedings of the IEEE conference on computer vision and pattern recognition}, pp.\  6154--6162, 2018{\natexlab{b}}.

\bibitem[Chen et~al.(2024{\natexlab{a}})Chen, Xu, Kirmani, Ichter, Sadigh, Guibas, and Xia]{vlmReason}
Boyuan Chen, Zhuo Xu, Sean Kirmani, Brain Ichter, Dorsa Sadigh, Leonidas Guibas, and Fei Xia.
\newblock Spatialvlm: Endowing vision-language models with spatial reasoning capabilities.
\newblock In \emph{Proceedings of the IEEE/CVF Conference on Computer Vision and Pattern Recognition}, pp.\  14455--14465, 2024{\natexlab{a}}.

\bibitem[Chen et~al.(2024{\natexlab{b}})Chen, Wu, Chitta, Jaeger, Geiger, and Li]{chen2024end}
Li~Chen, Penghao Wu, Kashyap Chitta, Bernhard Jaeger, Andreas Geiger, and Hongyang Li.
\newblock End-to-end autonomous driving: Challenges and frontiers.
\newblock \emph{IEEE Transactions on Pattern Analysis and Machine Intelligence}, 2024{\natexlab{b}}.

\bibitem[Cossu et~al.(2021)Cossu, Carta, Lomonaco, and Bacciu]{cl4survey2}
Andrea Cossu, Antonio Carta, Vincenzo Lomonaco, and Davide Bacciu.
\newblock Continual learning for recurrent neural networks: an empirical evaluation.
\newblock \emph{Neural Networks}, 143:\penalty0 607--627, 2021.

\bibitem[Daoud et~al.(2024)Daoud, Bunel, and Gu{\'e}riau]{cornersim123}
Alaa Daoud, Corentin Bunel, and Maxime Gu{\'e}riau.
\newblock Cornersim: A virtualization framework to generate realistic corner-case scenarios for autonomous driving perception testing.
\newblock \emph{Procedia Computer Science}, 238:\penalty0 184--191, 2024.

\bibitem[Geiger et~al.(2013)Geiger, Lenz, Stiller, and Urtasun]{kitti}
Andreas Geiger, Philip Lenz, Christoph Stiller, and Raquel Urtasun.
\newblock Vision meets robotics: The kitti dataset.
\newblock \emph{The International Journal of Robotics Research}, 32\penalty0 (11):\penalty0 1231--1237, 2013.

\bibitem[Greer \& Trivedi(2024)Greer and Trivedi]{vlm4human}
Ross Greer and Mohan Trivedi.
\newblock Towards explainable, safe autonomous driving with language embeddings for novelty identification and active learning: Framework and experimental analysis with real-world data sets.
\newblock \emph{arXiv preprint arXiv:2402.07320}, 2024.

\bibitem[Guo et~al.(2024)Guo, Lykov, Yagudin, Konenkov, and Tsetserukou]{vlm4codriver}
Ziang Guo, Artem Lykov, Zakhar Yagudin, Mikhail Konenkov, and Dzmitry Tsetserukou.
\newblock Co-driver: Vlm-based autonomous driving assistant with human-like behavior and understanding for complex road scenes.
\newblock \emph{arXiv preprint arXiv:2405.05885}, 2024.

\bibitem[Jiang et~al.(2020)Jiang, Xu, Zhang, Liang, and Li]{jiang2020sp}
Chenhan Jiang, Hang Xu, Wei Zhang, Xiaodan Liang, and Zhenguo Li.
\newblock Sp-nas: Serial-to-parallel backbone search for object detection.
\newblock In \emph{Proceedings of the IEEE/CVF conference on computer vision and pattern recognition}, pp.\  11863--11872, 2020.

\bibitem[Jin et~al.(2025)Jin, Feng, Mou, Decker, Lakemeyer, Simons, and Stegmaier]{vlmFeature}
Er~Jin, Qihui Feng, Yongli Mou, Stefan Decker, Gerhard Lakemeyer, Oliver Simons, and Johannes Stegmaier.
\newblock Logicad: Explainable anomaly detection via vlm-based text feature extraction.
\newblock \emph{arXiv preprint arXiv:2501.01767}, 2025.

\bibitem[Kirkpatrick et~al.(2017)Kirkpatrick, Pascanu, Rabinowitz, Veness, Desjardins, Rusu, Milan, Quan, Ramalho, Grabska-Barwinska, et~al.]{cl4jj1}
James Kirkpatrick, Razvan Pascanu, Neil Rabinowitz, Joel Veness, Guillaume Desjardins, Andrei~A Rusu, Kieran Milan, John Quan, Tiago Ramalho, Agnieszka Grabska-Barwinska, et~al.
\newblock Overcoming catastrophic forgetting in neural networks.
\newblock \emph{Proceedings of the national academy of sciences}, 114\penalty0 (13):\penalty0 3521--3526, 2017.

\bibitem[Kowol et~al.(2022)Kowol, Bracke, and Gottschalk]{cc4carla}
Kamil Kowol, Stefan Bracke, and Hanno Gottschalk.
\newblock A-eye: Driving with the eyes of ai for corner case generation.
\newblock \emph{arXiv preprint arXiv:2202.10803}, 2022.

\bibitem[Li et~al.(2022)Li, Chen, Wang, Hong, Ye, Han, Chen, Zhang, Xu, Yeung, et~al.]{coda123}
Kaican Li, Kai Chen, Haoyu Wang, Lanqing Hong, Chaoqiang Ye, Jianhua Han, Yukuai Chen, Wei Zhang, Chunjing Xu, Dit-Yan Yeung, et~al.
\newblock Coda: A real-world road corner case dataset for object detection in autonomous driving.
\newblock In \emph{European Conference on Computer Vision}, pp.\  406--423. Springer, 2022.

\bibitem[Liu et~al.(2021)Liu, Lin, Cao, Hu, Wei, Zhang, Lin, and Guo]{model5}
Ze~Liu, Yutong Lin, Yue Cao, Han Hu, Yixuan Wei, Zheng Zhang, Stephen Lin, and Baining Guo.
\newblock Swin transformer: Hierarchical vision transformer using shifted windows.
\newblock In \emph{Proceedings of the IEEE/CVF international conference on computer vision}, pp.\  10012--10022, 2021.

\bibitem[Lopez-Paz \& Ranzato(2017)Lopez-Paz and Ranzato]{gem}
David Lopez-Paz and Marc'Aurelio Ranzato.
\newblock Gradient episodic memory for continual learning.
\newblock \emph{Advances in neural information processing systems}, 30, 2017.

\bibitem[Mallya \& Lazebnik(2018)Mallya and Lazebnik]{packnet}
Arun Mallya and Svetlana Lazebnik.
\newblock Packnet: Adding multiple tasks to a single network by iterative pruning.
\newblock In \emph{Proceedings of the IEEE conference on Computer Vision and Pattern Recognition}, pp.\  7765--7773, 2018.

\bibitem[Mao et~al.(2021)Mao, Niu, Jiang, Liang, Chen, Liang, Li, Ye, Zhang, Li, et~al.]{mao2021one}
Jiageng Mao, Minzhe Niu, Chenhan Jiang, Hanxue Liang, Jingheng Chen, Xiaodan Liang, Yamin Li, Chaoqiang Ye, Wei Zhang, Zhenguo Li, et~al.
\newblock One million scenes for autonomous driving: Once dataset.
\newblock \emph{arXiv preprint arXiv:2106.11037}, 2021.

\bibitem[Pan et~al.(2024)Pan, Yaman, Nesti, Mallik, Allievi, Velipasalar, and Ren]{vlm4planning}
Chenbin Pan, Burhaneddin Yaman, Tommaso Nesti, Abhirup Mallik, Alessandro~G Allievi, Senem Velipasalar, and Liu Ren.
\newblock Vlp: Vision language planning for autonomous driving.
\newblock In \emph{Proceedings of the IEEE/CVF Conference on Computer Vision and Pattern Recognition}, pp.\  14760--14769, 2024.

\bibitem[Press(2024)]{tesla2024}
Associated Press.
\newblock Tesla’s ‘full self-driving’ under investigation after pedestrian killed.
\newblock {https://apnews.com/article/tesla-full-self-driving-investigation-} \\ {pedestrian-killed-f2121166d60d85bd173a734c91049e73}, 2024.

\bibitem[Ren et~al.(2016)Ren, He, Girshick, and Sun]{model2}
Shaoqing Ren, Kaiming He, Ross Girshick, and Jian Sun.
\newblock Faster r-cnn: Towards real-time object detection with region proposal networks.
\newblock \emph{IEEE transactions on pattern analysis and machine intelligence}, 39\penalty0 (6):\penalty0 1137--1149, 2016.

\bibitem[Ross \& Doll{\'a}r(2017)Ross and Doll{\'a}r]{model1}
T-YLPG Ross and GKHP Doll{\'a}r.
\newblock Focal loss for dense object detection.
\newblock In \emph{proceedings of the IEEE conference on computer vision and pattern recognition}, pp.\  2980--2988, 2017.

\bibitem[Rusu et~al.(2016)Rusu, Rabinowitz, Desjardins, Soyer, Kirkpatrick, Kavukcuoglu, Pascanu, and Hadsell]{pnn}
Andrei~A Rusu, Neil~C Rabinowitz, Guillaume Desjardins, Hubert Soyer, James Kirkpatrick, Koray Kavukcuoglu, Razvan Pascanu, and Raia Hadsell.
\newblock Progressive neural networks.
\newblock \emph{arXiv preprint arXiv:1606.04671}, 2016.

\bibitem[Sun et~al.(2021{\natexlab{a}})Sun, Feng, Yan, and Liu]{cc4make}
Haowei Sun, Shuo Feng, Xintao Yan, and Henry~X Liu.
\newblock Corner case generation and analysis for safety assessment of autonomous vehicles.
\newblock \emph{Transportation research record}, 2675\penalty0 (11):\penalty0 587--600, 2021{\natexlab{a}}.

\bibitem[Sun et~al.(2020)Sun, Kretzschmar, Dotiwalla, Chouard, Patnaik, Tsui, Guo, Zhou, Chai, Caine, et~al.]{waymo}
Pei Sun, Henrik Kretzschmar, Xerxes Dotiwalla, Aurelien Chouard, Vijaysai Patnaik, Paul Tsui, James Guo, Yin Zhou, Yuning Chai, Benjamin Caine, et~al.
\newblock Scalability in perception for autonomous driving: Waymo open dataset.
\newblock In \emph{Proceedings of the IEEE/CVF conference on computer vision and pattern recognition}, pp.\  2446--2454, 2020.

\bibitem[Sun et~al.(2021{\natexlab{b}})Sun, Zhang, Jiang, Kong, Xu, Zhan, Tomizuka, Li, Yuan, Wang, et~al.]{sun2021sparse}
Peize Sun, Rufeng Zhang, Yi~Jiang, Tao Kong, Chenfeng Xu, Wei Zhan, Masayoshi Tomizuka, Lei Li, Zehuan Yuan, Changhu Wang, et~al.
\newblock Sparse r-cnn: End-to-end object detection with learnable proposals.
\newblock In \emph{Proceedings of the IEEE/CVF conference on computer vision and pattern recognition}, pp.\  14454--14463, 2021{\natexlab{b}}.

\bibitem[Udupa et~al.(2024)Udupa, Gurunath, Sikdar, and Sundaram]{Sim2Real}
Sumanth Udupa, Prajwal Gurunath, Aniruddh Sikdar, and Suresh Sundaram.
\newblock Mrfp: Learning generalizable semantic segmentation from sim-2-real with multi-resolution feature perturbation.
\newblock In \emph{Proceedings of the IEEE/CVF Conference on Computer Vision and Pattern Recognition}, pp.\  5904--5914, 2024.

\bibitem[Wang et~al.(2024)Wang, Zhang, Su, and Zhu]{cl4survey}
Liyuan Wang, Xingxing Zhang, Hang Su, and Jun Zhu.
\newblock A comprehensive survey of continual learning: theory, method and application.
\newblock \emph{IEEE Transactions on Pattern Analysis and Machine Intelligence}, 2024.

\bibitem[Xu et~al.(2024)Xu, Hu, Zhang, Meyer, Mustikovela, Srinivasa, Wolff, and Huang]{vlm4techer}
Yi~Xu, Yuxin Hu, Zaiwei Zhang, Gregory~P Meyer, Siva~Karthik Mustikovela, Siddhartha Srinivasa, Eric~M Wolff, and Xin Huang.
\newblock Vlm-ad: End-to-end autonomous driving through vision-language model supervision.
\newblock \emph{arXiv preprint arXiv:2412.14446}, 2024.

\bibitem[Zhao et~al.(2024{\natexlab{a}})Zhao, Duan, Su, Guo, Chen, Luo, et~al.]{coda_track2}
Jiawei Zhao, Yiting Duan, Jinming Su, Tingyi Guo, Xingyue Chen, Junfeng Luo, et~al.
\newblock Loop mining large-scale unlabeled data for corner case detection in autonomous driving.
\newblock In \emph{ECCV 2024 Workshop on Multimodal Perception and Comprehension of Corner Cases in Autonomous Driving}, 2024{\natexlab{a}}.

\bibitem[Zhao et~al.(2024{\natexlab{b}})Zhao, Zhao, Deng, Wang, Zhang, Zheng, Cao, Nan, Lian, and Burke]{zhao2024autonomous}
Jingyuan Zhao, Wenyi Zhao, Bo~Deng, Zhenghong Wang, Feng Zhang, Wenxiang Zheng, Wanke Cao, Jinrui Nan, Yubo Lian, and Andrew~F Burke.
\newblock Autonomous driving system: A comprehensive survey.
\newblock \emph{Expert Systems with Applications}, 242:\penalty0 122836, 2024{\natexlab{b}}.

\bibitem[Zhong et~al.(2024)Zhong, Liao, Zhang, Zhang, and Wang]{vlmClass}
Lanfeng Zhong, Xin Liao, Shaoting Zhang, Xiaofan Zhang, and Guotai Wang.
\newblock Vlm-cpl: Consensus pseudo labels from vision-language models for human annotation-free pathological image classification.
\newblock \emph{arXiv preprint arXiv:2403.15836}, 2024.

\bibitem[Zhu et~al.(2020{\natexlab{a}})Zhu, Su, Lu, Li, Wang, Dai, and DETR]{zhu2020deformable}
X~Zhu, W~Su, L~Lu, B~Li, X~Wang, J~Dai, and Deformable DETR.
\newblock deformable transformers for end-to-end object detection.
\newblock \emph{URL: https://arxiv. org/abs/2010.04159 (Accessed 29.11. 2023)}, 2020{\natexlab{a}}.

\bibitem[Zhu et~al.(2020{\natexlab{b}})Zhu, Su, Lu, Li, Wang, and Dai]{model4}
Xizhou Zhu, Weijie Su, Lewei Lu, Bin Li, Xiaogang Wang, and Jifeng Dai.
\newblock Deformable detr: Deformable transformers for end-to-end object detection.
\newblock \emph{arXiv preprint arXiv:2010.04159}, 2020{\natexlab{b}}.

\bibitem[Zuo et~al.(2025)Zuo, Hu, Zhou, Cui, Liu, Wang, Guan, Wang, and Xue]{ralad}
Jiacheng Zuo, Haibo Hu, Zikang Zhou, Yufei Cui, Ziquan Liu, Jianping Wang, Nan Guan, Jin Wang, and Chun~Jason Xue.
\newblock Ralad: Bridging the real-to-sim domain gap in autonomous driving with retrieval-augmented learning, 2025.
\newblock URL \url{https://arxiv.org/abs/2501.12296}.

\end{thebibliography}
\bibliographystyle{colm2025_conference}


\end{document}